\documentclass{article}
\usepackage{spconf,amsmath,graphicx}
\usepackage{amssymb}
\usepackage{multirow}
\usepackage{color, soul}
\usepackage{tabularx,booktabs}
\usepackage{subcaption}
\let\vec\mathbf
\newcommand\blfootnote[1]{%
  \begingroup
  \renewcommand\thefootnote{}\footnote{#1}%
  \addtocounter{footnote}{-1}%
  \endgroup
}

\title{Centroid-based Deep Metric Learning For Speaker Recognition}
\name{Jixuan Wang$^{1,2}$, Kuan-Chieh Wang$^{1,2}$, Marc T. Law$^{1,2}$, Frank Rudzicz$^{1,2,3,4}$, Michael Brudno$^{1,2,5}$}
\address{$^{1}$University of Toronto, Canada ~ $^{2}$Vector Institute, Canada ~ $^{3}$St Michael’s Hospital, Canada \\ $^{4}$Surgical Safety Technologies Inc, Canada ~ $^{5}$ Hospital for Sick Children, Canada \\
\{jixuan, wangkua1, law, frank, brudno\}@cs.toronto.edu}
\begin{document}
%\ninept
%
\maketitle
\begin{abstract}
\blfootnote{The first two authors contributed equally to this work.}Speaker embedding models that utilize neural networks to map utterances to a space where distances reflect similarity between speakers have driven recent progress in the speaker recognition task. However, there is still a significant performance gap between recognizing speakers in the training set and unseen speakers.  The latter case corresponds to the few-shot learning task, where a trained model is evaluated on unseen classes. Here, we optimize a speaker embedding model with \textit{prototypical network loss} (PNL), a state-of-the-art approach for the few-shot image classification task.  
The resulting embedding model outperforms the state-of-the-art \textit{triplet loss} based models in both speaker verification and identification tasks, for both seen and unseen speakers.   

%\JX{replaced this: }The resulting embedding model obtains state-of-the-art performance in both speaker verification and identification tasks, for both seen and unseen speakers.

\end{abstract}
\begin{keywords}
deep metric learning, triplet loss, sequence embedding, speaker recognition
\end{keywords}
\section{Introduction}
\label{sec:intro}

The development of embedding models to represent speech features in high-dimensional space has enabled elegant solutions to speaker recognition problems, including speaker verification (SV), and speaker identification (SI). Speaker verification systems verify whether or not a given utterance comes from some claimed speaker, while only having access to a handful of enrolled utterances (i.e. training examples). In the speaker identification task, a trained model is asked to classify among $K$ speakers given small amount of enrollment speech. For both tasks, finding good representation of speech features is essential to recognition of the speaker, especially with small training sets: the variation in phrases needs to be normalized, while variation across speakers must be preserved.

Traditional pipelines that combine i-vector and probabilistic linear discriminant analysis (PLDA) separately train the feature extractor and the final classifier~\cite{dehak2011front}.  However, performance of i-vector systems drops for short speech utterances ~\cite{sarkar2012study}. More robust feature extractors have been proposed, including replacing i-vectors with features extracted from deep networks~\cite{snyder2018x,snyder2017deep}.
%  These have been shown to outperform the i-vector approaches~\cite{heigold2016end, snyder2017deep, snyder2018x, li2017deep}.
Most recent efforts have relied on optimizing an end-to-end deep model with the triplet loss (TL)~\cite{li2017deep,zhang2017end,bredin2017tristounet,chung2018voxceleb2} and the generalized end-to-end (GE2E) loss~\cite{wan2018generalized} to build the speaker embeddings . 
%\KCW{repeated sentence}
 %Note we refer speaker recognition as the collection speaker-related tasks including SV and SI.

In this paper we propose the use of prototypical network loss (PNL) to optimize an end-to-end speaker embedding network. PNL was introduced for the few-shot image classification task \cite{snell2017prototypical} and is the state-of-the-art approach on several few-shot learning benchmarks. However, to the best of our knowledge, PNL has not been applied to speaker embedding or related problems. Here we show that for SV and SI tasks, a model trained with PNL outperforms an embedding network of the same architecture optimized with TL.
% Our work also compares different learning methods for deep embedding models. 
We discuss why PNL is a better formulation for learning an embedding model and provide empirical observations as to why it might be easier to use in practice.

\begin{figure}[!t]

\begin{minipage}[b]{.48\linewidth}
  \centering
  \centerline{\includegraphics[width=4.3cm]{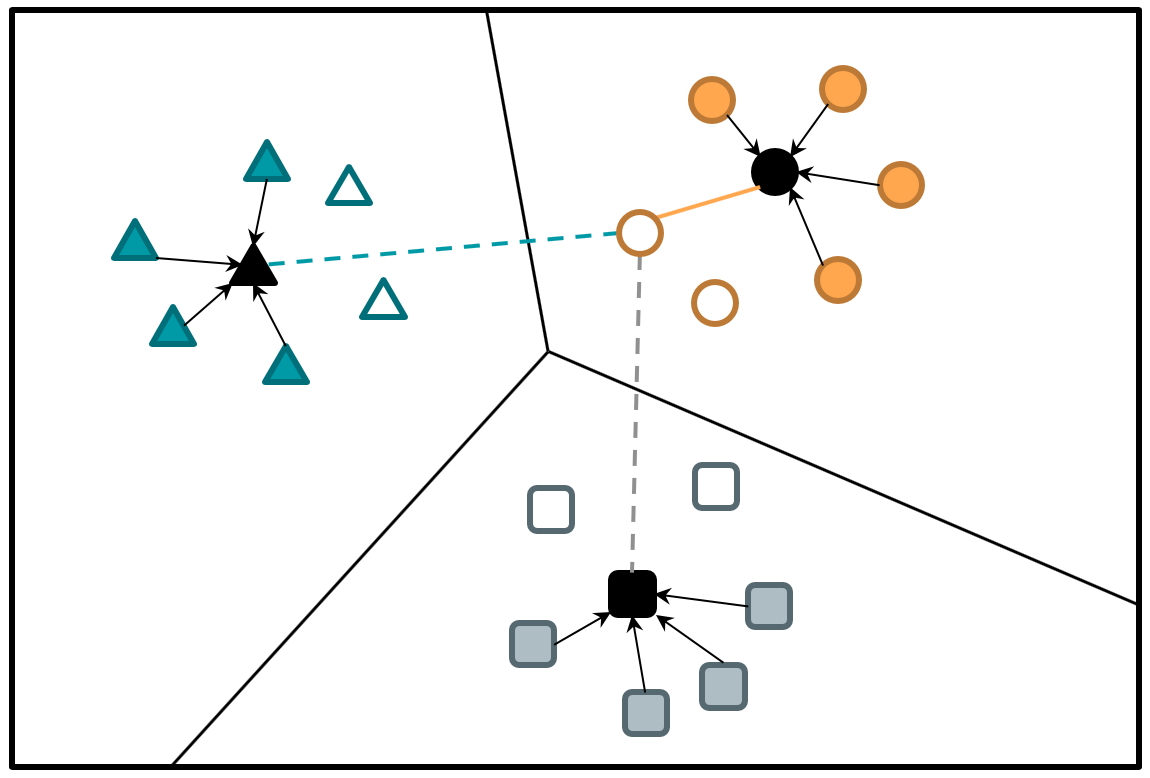}}
%  \vspace{1.5cm}
  \centerline{(a) prototypical network loss}\medskip
  \label{fig:fewshot}
\end{minipage}
\hfill
\begin{minipage}[b]{0.48\linewidth}
  \centering
  \centerline{\includegraphics[width=4.3cm]{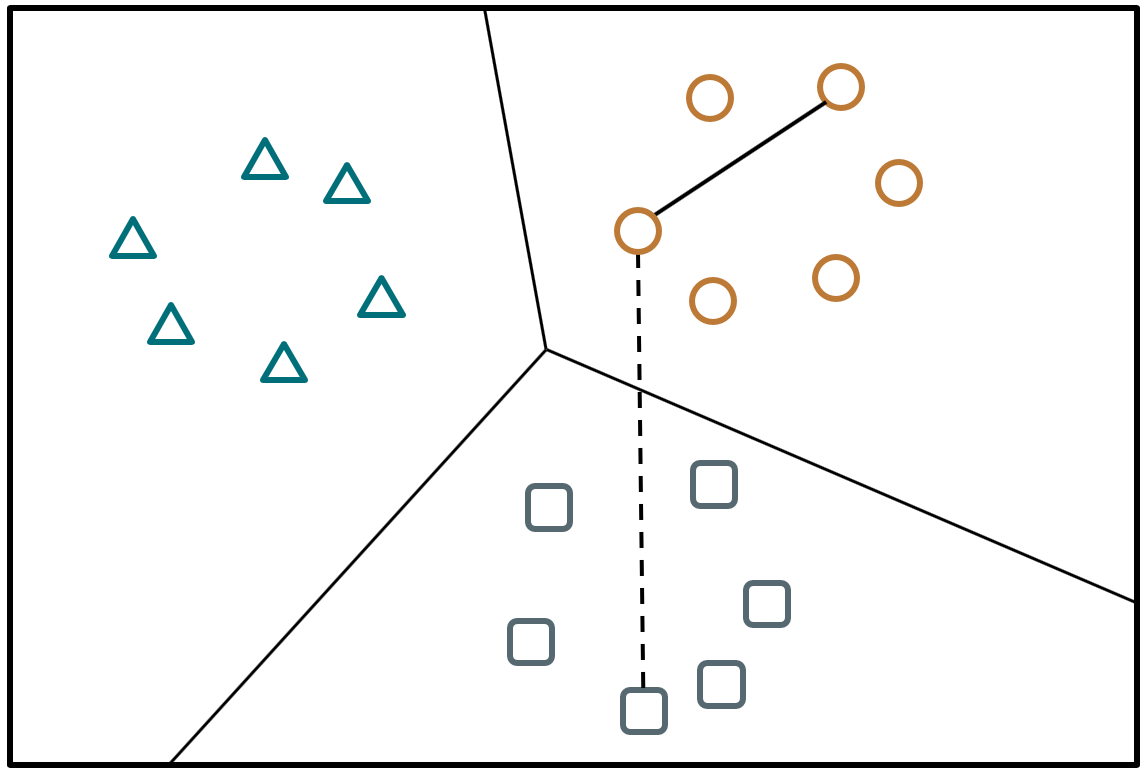}}
%  \vspace{1.5cm}
  \centerline{(b) triplet loss}\medskip
  \label{triplet}
\end{minipage}
\vspace{-1em}
\caption{
Comparison between the effect of prototypical network loss (PNL) and triplet loss (TL) on the embedding. Dashed lines represent distances encouraged to increase, while solid lines represent distances being decreased. \textbf{left}: For PNL, prototypes for different speakers, denoted by black nodes are computed as the mean of the support set (shaded) during training. \textbf{right}: 
For TL, a triplet consists of an anchor, positive, and negative samples, forming the (anchor, positive) and (anchor,negative) pairs. Depending on the sampling strategy not all triplets may be considered.
%\JX{replace: Anchor sets are selected randomly, so distance to representatives for same/different speakers will vary, and some triplets will contribute little or no information.}
}
\label{fig:compare}
\vspace{-1em}
\end{figure}

% Few-shot learning~\cite{fei2006one,lake2015human} is the task of performing classification on classes give a handful of labeled examples during test time. It is quite similar to many use cases of speaker verification and speaker identification. 
% For instance, the speakers encountered during test time are not necessarily seen during training. It is then essential to build profiles for unseen speakers with limited amount of data. 
% However, to the best of our knowledge, few-shot learning approaches have not been studied in this domain. In this paper, we propose to use few-shot learning approach to learn speaker embeddings. With the same speech sequence embedding models trained identical data set, we empirically show that few-shot approach outperforms the triplet loss based models and demonstrate potential for speaker verification and identification tasks.

\section{Related work}
\label{sec:relatedwork}

\textbf{Speaker embedding networks:} 
In traditional i-vector based methods for speaker embedding, a universal background model is first built. Then, a PLDA model is trained to measure the similarity of i-vectors. Replacing traditional i-vectors with speaker embedding models based on deep neural networks has lead to improvement in SV \cite{snyder2017deep, snyder2018x}. Nonetheless, a PLDA classifier is still needed to compare the similarity of embeddings. More recently,
% end-to-end approaches optimizing the TL were investigated~\cite{zhang2017end, li2017deep, bredin2017tristounet, chung2018voxceleb2}. 
end-to-end training of an embedding network that makes decision by comparing distance in the embedding to a cross-validated threshold outperformed traditional methods.  For detailed comparison between embedding networks and i-vector based methods, we refer the reader to \cite{zhang2017end,snyder2017deep, snyder2018x}. %\KCW{make sure no citations missing}.
Building on top of these studies, our work focuses on the comparison between two different approaches for deep metric learning (TL ~\cite{li2017deep,zhang2017end,bredin2017tristounet,chung2018voxceleb2}  and PNL \cite{snell2017prototypical}) for end-to-end speaker embedding models. 

\textbf{Deep metric learning:} 
End-to-end speaker embedding models can be seen as a form of deep metric learning, which has been widely studied in the machine learning literature. 
Early examples of metric learning with neural networks include signature \cite{bromley1994signature} and face verification \cite{chopra2005learning}. Both compare pairs of examples with standard similarity functions (e.g. cosine or Euclidean distance) at the final embedding layer of a \textit{siamese} architecture. % (i.e., a network which processes two inputs using the same set of parameters)
More complex loss functions involving \textit{triplets} \{\textit{anchor}, \textit{positive}, \textit{negative}\} were later proposed \cite{schultz2004learning}, and shown to perform well on face verification~\cite{schroff2015facenet}. %, and the learned embedding can lead to better classification and retrieval performance \cite{weinberger2009distance, oh2016deep}. 
% In the speech recognition literature, the TL (or contrastive loss referred in \cite{chung2018voxceleb2}) is the method of choice for speaker embedding~\cite{li2017deep, bredin2017tristounet, zhang2017end}.

\textbf{Few-shot learning:} 
%Few-shot learning~\cite{fei2006one,lake2015human} is the task of performing classification on classes when given only a few labeled examples at test time. 
Motivated by the fact that humans can learn new concepts from only a handful of examples, researchers have proposed the challenging task of ``few-shot learning'' ~\cite{fei2006one,lake2015human}. 
The test-time task is to classify examples among $K$ new classes (i.e. unseen during training) while only being given a handful of labeled examples from these new classes. 
The same consideration arises naturally for speaker recognition tasks, as speakers encountered during test time may be different. It is essential to build profiles for previously unseen speakers with limited data. 
In contrast to previous applications of PNL \cite{snell2017prototypical}, we use it in conjunction with a recurrent neural network for sequential data (i.e. speech) rather than a convolutional network for images.
%\textit{Episodic training} refers to constructing training mini-batches that emulate the test-time tasks \cite{snell2017prototypical,vinyals2016matching}. 
%and earlier attempts using siamese networks without episodic training \cite{koch2015siamese}. 
%Few-shot learning is analogous to some speaker recognition tasks as both entail learning a good representation of training data that is able to generalize to classes unseen during training (e.g., new speakers). 

% Prototypical networks have been used in few-shot image classification setting with convolutional networks \cite{snell2017prototypical}. Each category is represented by a prototype calculated from labeled examples, and loss (PNL) is defined as distance from the prototype. Test queries are assigned to the category with the closest prototype.  \KCW{I don't think we need to say this here. It's well described in Sec. 3.2}

% We apply PNL to features extracted from utterances through a recurrent network. \KCW{this sounds like the RNN is trained as a feature extractor, on top of which we apply PNL}

% Our work focuses on the comparison between different learning methods for deep embedding models.  

\section{Optimization schemes \& Model}

We now explain the standard triplet loss scheme and compare it to PNL \cite{snell2017prototypical}. We then describe the model we optimize using the two schemes for speaker embedding.

\vspace{-0.5em}
\label{sec:fewshot}
\subsection{Triplet loss}

For triplet-based models, we denote by $S' = \{\vec{x}_1, \cdots, \vec{x}_{N'}\}$ the examples in one mini-batch of size $N'$, where $\vec{x}_i$ is a sequence of speech features. These models sample triplets, which consist of an anchor $\vec{x}_a$, a positive sample $\vec{x}_p$ with the same speaker label, and a negative sample $\vec{x}_n$ with a different speaker label.
For each triplet $\tau = (\vec{x}_a, \vec{x}_p, \vec{x}_n)$, the triplet loss is formulated as: 
\begin{equation}
L(\tau) = max(0, d_{a,p} - d_{a,n} +\alpha)
\end{equation}
where $d_{a,b} = d(f(\vec{x}_a), f(\vec{x}_b))$, $d$ and $f$ are the distance function (e.g. cosine or squared Euclidean distance) and speaker embedding model, respectively, and $\alpha > 0$ is a margin. Minimizing $L(\vec{x}_a, \vec{x}_p, \vec{x}_n)$ learns representations so that the similar pair $(\vec{x}_a, \vec{x}_p)$ has smaller distance than the dissimilar pair $(\vec{x}_a, \vec{x}_n)$, adjusted by the margin. It is worth noting that the triplet loss does not minimize distances between similar pairs (i.e., when $d_{a,p} - d_{a,n}+\alpha <0 $); it only tries to preserve some order between distances. 
Finally, the loss for a mini-batch is 
%\begin{equation}
    $J_{TL} = \sum_{\tau \in \mathrm{T} } L (\tau)$ 
%\end{equation}
where $\mathrm{T}$ is the set all of possible triplets in the mini-batch. 
In the context of very large datasets, creating all the possible triplets is computationally expensive; different triplet sampling strategies have been proposed to ensure fast convergence while avoiding degenerate solutions. 
For instance, the semi-hard mining strategy \cite{schroff2015facenet} samples only one hard negative pair for each positive pair. A triplet is ``hard" if $d_{a,p} - d_{a,n}+\alpha > 0$. 
%As showed in \JX{[]}, converge speed and generalization ability of models trained on triplet loss can be influenced by the choice of triplet sampling strategy. 
%Other than the na\"ive sampling strategy, which utilizes all possible triplets in each batch, another commonly used strategy is called semi-hard \JX{paper and name}. 
%According to this strategy, only one negative pair is sampled for each positive pair. Also, the negative sample selected should be ``hard", which means . In other words, we want to guarantee the loss of each triplet should not be zero so that it can contribute to the model optimization. 
%However, no matter which triplet sampling strategy is used, there still might be some triplets of which the loss can not be optimized, where samples belonging to the same class are split by negative samples into sub-clusters~\cite{song2017deep}. As a result, the average loss of triplet might increase with training.  

\vspace{-0.5em}
\subsection{Prototypical Networks Loss} 
\label{sec:protonet}
      
Prototypical Networks \cite{snell2017prototypical} train a neural network episodically; each episode is composed of one mini-batch containing $K$ categories (here, speakers). 
The mini-batch contains a support set called $S$ and a query set called $Q$. 
In our case, 
the support set $S = \{(\vec{x}_i, y_i)\}_{i=1}^N$ represents each example as a sequence of speech feature vectors $\vec{x}_i$ with corresponding speaker label $y_i \in \{ 1, \cdots, K \}$. We denote $S_k \subseteq S$ as the set of examples in $S$ of speaker $k$.

The prototype (or centroid) of each class $\vec{c}_k \in \mathbb{R}^M $ is calculated as the mean of embeddings in the support set:
\begin{equation}
    \vec{c}_k = \frac{1}{|S_k|} \sum_{(\vec{x}_i, y_i) \in S_k}  f(\vec{x}_i)
\end{equation}
where $f$ is the speaker embedding model which maps sequences of speech features into the $M$-dimensional embedding space (see details in Section \ref{sec:f}).

During training, each query example $\{ ( \vec{x}_j, y_j ) \} \in Q$ is classified against $K$ speakers based on a softmax over distances to each speaker prototypes: 
%\FR{softmax is well-known; I'd keep this unless space becomes an issure, then remove}:
\begin{equation}
    p (y = y_j | \vec{x}_j) = \frac{\exp \left( -d(f(\vec{x}_j), \vec{c}_{y_j}) \right)}{\Sigma_{k'} \exp \left(-d(f(\vec{x}_j), \vec{c}_{k'} \right)}
\end{equation}
where $d$ is the distance function. 
%The loss is the negative log-probability of the true label: $-\log p (y = y_j | \vec{x}_j)$. 
The loss function for each mini-batch is
%\begin{equation}
    $J_{PNL} = \sum_{\{ ( \vec{x}_j, y_j ) \} \in Q } - \log p(y = y_j | \vec{x}_j)$
%\end{equation}
% PNL can be seen as a supervised clustering method \cite{anonymous2019dimensionality} where $f$ is learned so that applying a clustering algorithm on the learned representations returns the desired partitions. 

\subsection{Speech sequence embedding model} \label{sec:f}
\begin{figure}[!t]

\begin{minipage}[b]{1.0\linewidth}
  \centering
  \centerline{\includegraphics[width=7.5cm]{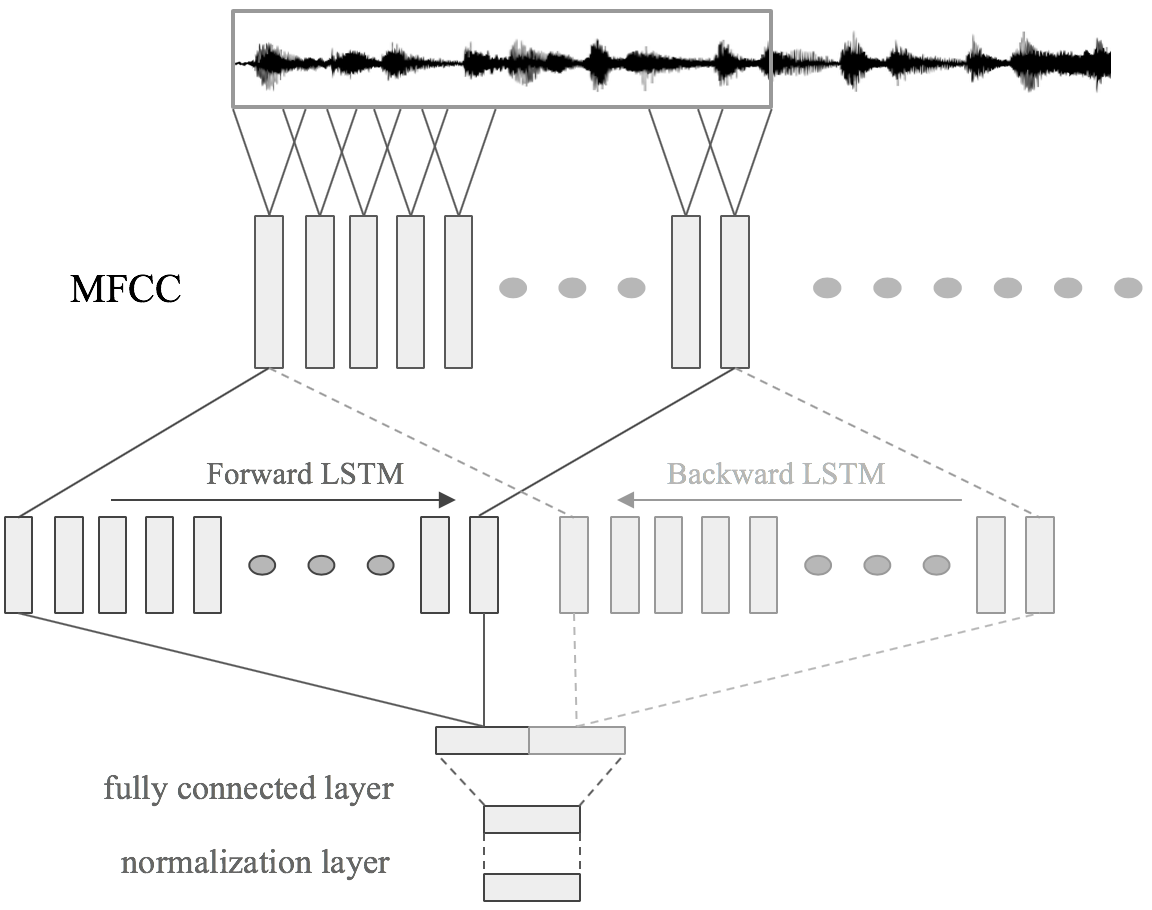}}
%  \vspace{1.5cm}
%  \centerline{(a) Prototypical Networks}\medskip
  \label{fig:fewshot}
\end{minipage}
\caption{Speech sequence embedding model}
\label{fig:speechemb}
\end{figure}

In terms of network architecture, we use the same speech sequence embedding model as TristouNet~\cite{bredin2017tristounet}. We use the same model architecture when optimizing with each of the two losses above. As shown in Fig.~\ref{fig:speechemb}, the sequence of Mel-frequency cepstral coefficients (MFCC) features are collected. They are then fed into bidirectional LSTM~\cite{hochreiter1997long}. 
The outputs from the forward and backward LSTMs are first average-pooled over time, concatenated, then processed by a fully connected layer and a normalization layer.

\section{Experiment}
\label{sec:experiment}
In this section, we introduce the experiments conducted for evaluation and comparison of different models. 

\subsection{Datasets and implementation details}
We use the VCTK corpus~\cite{veaux2016superseded} and a subset of VoxCeleb2 dataset~\cite{chung2018voxceleb2}. 
VCTK corpus contains clearly read speech, while VoxCeleb2 has more background noise and overlapping speech.
Speech data of the first 90 speakers in VCTK corpus was divided into training, validation and test sets.
Data of the remaining 18 speakers was used as an ``unseen'' set to evaluate the generalizability of the method. 
For VoxCeleb2, we selected a subset containing 101 speakers (we refer to this subset as VoxCeleb2 dataset for conciseness) and use data of 71 speakers for training and validation, while the other 30 speakers are used as the ``unseen'' set.

%\subsection{Implementation details}
We use the default implementation of TristouNet for feature extraction and speech sequence embedding models. In the following, we refer to the models by the loss used during optimization (i.e. PNL vs. TL) as the architecture of the embedding network is fixed. 

\textbf{Feature extraction} We extracted 19-dimensional MFCCs, their first and second derivatives, along with the first and second derivatives of energy in a 25ms window every 10ms using the \textit{pyannote} multimedia processing toolkit\footnote{http://pyannote.github.io/} and Yaafe toolkit~\cite{mathieu2010yaafe}. This results in 59-dimensional acoustic features. We use 2-second segments for both PNL and TL models to ensure comparability of results and to test performance on shorter utterances.

%As shown in~\cite{bredin2017tristounet}, speech sequence duration used for training can influence the model performance.

\textbf{Training} We use PyTorch~\cite{paszke2017automatic} for implementation of both losses\footnote{Our implementation of PNL is based on that of the original authors:  https://github.com/jakesnell/prototypical-networks \cite{snell2017prototypical}}. The output dimension is 16, and the $tanh$ activation function is used for the fully connected layer. Squared Euclidean distance is used as the distance function.
The margin $\alpha$ used in our implementation of TL is $0.2$ following \cite{bredin2017tristounet}.
For both models, the Adam optimizer~\cite{kingma2014adam} is used with $10^{-3}$ learning rate.
Both models are trained for 100 epochs. 

For each mini-batch, we randomly choose 15 speakers without replacement on VCTK (10 speakers on VoxCeleb2). We enforce equal mini-batch size between the two formulations (i.e., $|S|+|Q|=N'$). 
TL models are trained with Euclidian distance or cosine distance using na\"ive or semi-hard strategy. 
PNL models are trained with different size of support and query sets. 
To avoid confusion in the following, ``TL ($s$, $d$)" denotes sampling strategy $s$ with distance metric $d$, and ``PNL ($x$\_$y$, $d$)" denotes PNL using $x$-shot with query set of size $y$ training episodes.

\vspace{-1em}
\subsection{``\textit{Same/Different}" experiments}

\begin{figure}[!t]
\begin{minipage}[b]{1.0\linewidth}
  \centering
  \centerline{\includegraphics[width=9.0cm]{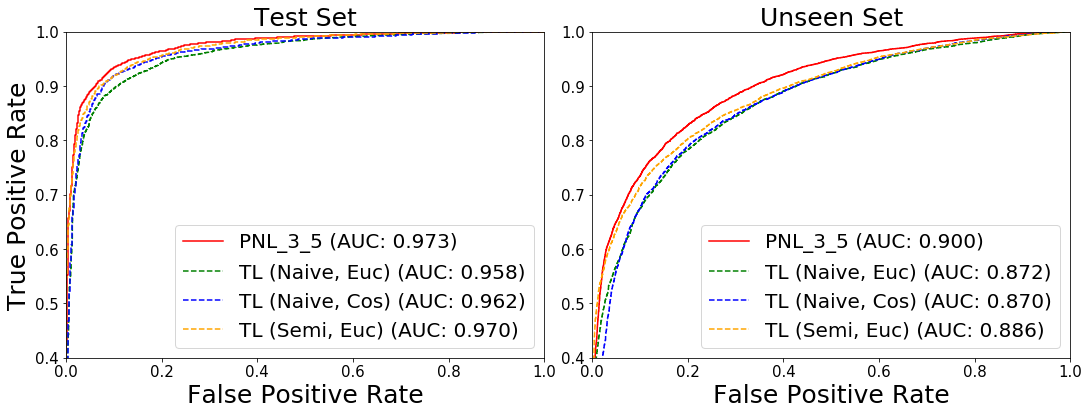}}
\end{minipage}
\caption{``\textit{same/different}" experiments on test and unseen sets.}
\label{fig:samediff}
\vspace{-1em}
\end{figure}

As with TristouNet~\cite{bredin2017tristounet}, we first conduct ``\textit{same/different}" experiments on the VCTK corpus. The same number of positive (same speaker) and negative (different speakers) pairs are randomly selected. 
For each pair of segments, we calculate the distance between their embeddings and compare it with a threshold to predict whether they are from the same or different speakers.

We report the receiver operating characteristic (ROC) curve for each model, shown in Fig.~\ref{fig:samediff}.  
All models perform comparably well on seen data set. However, on the unseen data PNL model outperforms TL models. 

\vspace{-1em}
\subsection{Speaker Identification}
% Our first experiment is with regards SI to multiclass classification 
% on batches. As in the training process, during testing, we randomly sample $K$ speakers, and each with $N$ speech segments. 
% For models trained with PNL, the classification accuracy is the mean of the accuracies of samples in that batch being classified as the corresponding speakers. The final result is the mean of classification accuracies of all batches.
% For models trained with TL, we use a similar evaluation strategy, i.e., randomly selecting several segments (as with the size of the query set in few-shot learning) for classification and the remained segments are used for enrollment. 

% This experiment is actually performing speaker identification. In each batch, speech segments in support set serve as enrollment data for speakers, based on which speaker embeddings are extracted and averaged for each speaker.
% Then the embeddings in query set are assigned to speakers according to their closest speaker embeddings. 
% The final result on each data set is the average classification accuracy across all batches.
% We conduct two set of experiments: 6-way (speaker) and 18-way classification, both on test and unseen sets.

To evaluate SI performance, we simulate each test \textit{task} with a batch of $K$ speakers, each with $S$ enrollment samples, and $Q$ query samples.  Classification of a query is done by finding the closest prototype based on some metric $d$.  
Results are shown in Table~\ref{tbl:siacc_vctk}, from which one can observe that the 3-shot PNL-based model outperforms the TL (Semi, Euc) model on all tasks, especially on the more challenging 18-way SI task (6\% and 19\% relative improvement on test and unseen data set, respectively). 
Interestingly, the one-shot PNL based model performs better than triplet loss based model with na\"ive sampling strategy while it ``sees" many fewer positive and negative pairs for each batch. It also performs nearly as well as TL that uses the more complicated semi-hard sampling strategy. 
% This demonstrates the superior generalization ability of PNL-based models. \KCW{if we want to keep this, be more specific}

\begin{table}[!t]
\caption{SI accuracy on test and unseen sets of VCTK. Under the Model column are the training configurations. The top row `S:$N_S$, Q:$N_Q$, $K$-way' denotes the \textit{task} configurations.}
\label{tbl:siacc_vctk}
\centering
\small
\begin{tabular}{ c | p{1cm} | p{1cm}| p{1cm}| p{1cm} } 
%\begin{tabularx}{\textwidth}{ |c| *{4}{Y|} }
\hline
\multirow{2}{2.5em}{Model} & \multicolumn{2}{c|}{S: 5, Q: 5, 6-way} & \multicolumn{2}{c}{S: 10, Q: 10, 18-way} \\\cline{2-5}

  & Test & Unseen & Test & Unseen \\
\hline
TL (Naive, Euc) & 91.96\% & 77.80\%  & 81.25\% & 56.37\% \\
TL (Naive, Cos) & 92.37\% & 77.69\% & 83.51\% & 58.51\%\\
TL (Semi, Euc) & 93.33\%  & 79.69\% & 85.38\% & 58.49\% \\
TL (Semi, Cos) & 92.13\%  & 73.94\% & 83.99\% & 52.97\% \\
\hline
PNL (1\_5, Euc) & 93.24\% & 77.90\% & 83.71\% & 56.52\% \\
\textbf{PNL (3\_5, Euc)} & \textbf{95.63\%} & 84.81\% & \textbf{90.53\%} & \textbf{69.64\%} \\
PNL (5\_5, Euc) & 95.47\% & 83.69\% & 89.85\% & 68.55\%\\
PNL (10\_10, Euc) & 94.38\% & \textbf{85.00\%} & 88.43\% & 66.63\% \\
%Triplet loss & 10 & 75.88\% \\
\hline
%\multirow{3}{10em}{Few/Zero-shot learning} & 5+5 & \textbf{83.15\%} \\ 

\end{tabular}
\end{table}

\begin{table}[!t]
\caption{SI accuracy on VoxCeleb2. 
%$S$ refers to number of 2-second segments used for enrollment, i.e ``$S$: 30" means we use 1 minute of speech as enrollment for each speaker. ``$Q$: -" means speaker identification is conducted on the remaining data of each speaker.
}
\label{tbl:siacc_vox}
\centering
\small
\begin{tabular}{ c |  p{1cm}| p{1cm} | p{1cm} | p{1cm} } 
%\begin{tabularx}{\textwidth}{ |c| *{4}{Y|} }
\hline
\multirow{3}{2.5em}{Model}  & \multicolumn{4}{c}{15-way} \\\cline{2-5}
   & \multicolumn{2}{c|}{S: 10, Q: -} & \multicolumn{2}{c}{S: 30, Q: -} \\\cline{2-5}
  & Test & Unseen & Test & Unseen\\
\hline
TL (Semi, Cos) &  74.74\% & 53.92\% & 75.18\%  & 59.61\%\\
TL (Semi, Euc) &  71.78\% & 51.74\% & 72.02\% & 56.79\%\\
% \hline
% PNL (1\_5, Euc)  & 73.04\% & 52.85\% & 73.02\% & 57.81\%\\
% PNL (3\_5, Euc) & 77.39\% & 56.69\% & 78.15\% & 63.59\%\\
\textbf{PNL (5\_5, Euc)} &  \textbf{78.38\%} & \textbf{59.44\%} & \textbf{79.23\%} & \textbf{66.63\%} \\
%Triplet loss & 10 & 75.88\% \\
\hline
\end{tabular}
\end{table}
\vspace{-1em}

\begin{table}[!t]
\caption{EER of SV on both data sets. ``2s''  (2 seconds) refers the duration of speech we used for enrollment. }
\label{tbl:sv}
\begin{subtable}[h]{0.5\textwidth}
\centering
\vspace{-.5em}
\caption{EER on VCTK}
\label{tbl:eervctk}
\centering
\small
\begin{tabular}{ c |  p{1.6cm}| p{1.6cm} p{1.6cm} } 
%\begin{tabularx}{\textwidth}{ |c| *{4}{Y|} }
\hline
 \multirow{2}{2.5em}{Model}  & \multicolumn{1}{c|}{Test} & \multicolumn{2}{c}{Unseen} \\\cline{2-4}
  & 60s & 60s & 10s \\
\hline
TL (Semi, Cos)  &   5.43($\pm$0.16)  &   13.87($\pm$0.37) &   16.19($\pm$0.86) \\%&   22.37($\pm$1.96)  \\
TL (Semi, Euc)  &   5.05($\pm$0.09)  &   12.26($\pm$0.69) &   13.44($\pm$0.91) \\%&   19.53($\pm$1.64)  \\
\textbf{PNL (5\_5, Euc)} &   \textbf{4.08($\pm$0.13)}  &   \textbf{10.77($\pm$0.58)} &   \textbf{12.00($\pm$0.76)} \\%&   19.27($\pm$1.93)  \\
%Triplet loss & 10 & 75.88\% \\
\hline
%\multirow{3}{10em}{Few/Zero-shot learning} & 5+5 & \textbf{83.15\%} \\ 
\end{tabular}
\end{subtable}
\newline
\vspace*{0.3 cm}
\newline
\begin{subtable}[h]{0.5\textwidth}
\centering
\vspace{-.5em}
\caption{EER on VoxCeleb2}
\label{tbl:eervox}
\centering
\small
\begin{tabular}{ c |  p{1.6cm}| p{1.6cm} p{1.6cm} }  
%\begin{tabularx}{\textwidth}{ |c| *{4}{Y|} }
\hline
 \multirow{2}{2.5em}{Model}  & \multicolumn{1}{c|}{Test} & \multicolumn{2}{c}{Unseen} \\\cline{2-4}
  & 60s & 60s & 10s \\
\hline
TL (Semi, Cos)  &   9.23($\pm$0.13)  &   14.62($\pm$0.35) &   16.93($\pm$0.45) \\% &   23.02($\pm$0.80)  \\
TL (Semi, Euc)  &   9.90($\pm$0.11)  &   15.92($\pm$0.32) &   17.61($\pm$0.51) \\%&   22.85($\pm$1.14)  \\
\textbf{PNL (5\_5, Euc)} &   \textbf{8.29($\pm$0.12)}  &   \textbf{13.68($\pm$0.26)} &   \textbf{15.67($\pm$0.56)} \\%&   22.66($\pm$0.88)  \\
%Triplet loss & 10 & 75.88\% \\
\hline
%\multirow{3}{10em}{Few/Zero-shot learning} & 5+5 & \textbf{83.15\%} \\ 
\end{tabular}
\end{subtable}
\vspace{-1em}
\end{table}

\subsection{Speaker verification}
To evaluate SV performance, we randomly select some speech segments for enrollment. 
Then, 200 (resp. 100) segments of each speaker are selected as positive samples on VoxCeleb2 (resp. VCTK). Equal number of negative samples are selected from different speakers.
During enrollment phase, speaker prototypes are computed from the enrollment set.
For verification, the decision is made by comparing the distance between the embedding of the query segment and the speaker prototype to a threshold.
Performance is evaluated by equal error rate (EER). Results are shown in Table~\ref{tbl:sv}.

Results obtained by repeating the experiments 10 times (i.e. mean and standard deviation) on VCTK and VoxCeleb2 are shown in Table~\ref{tbl:eervctk} and Table~\ref{tbl:eervox}, respectively.
EER of PNL model is significantly lower than that of TL models ($p \ll 0.001$ using t-test) across all tasks.
As expected, both models perform reasonably well on test set of VCTK, while a little worse on test set of VoxCeleb2.   
For unseen set, EERs of all models decrease for longer duration of speech data for enrollment.
Although TL (Semi, Cos) outperforms TL (Semi, Euc) on more noisy data set, PNL still achieves the lowest EER.

\vspace{-.5em}
\subsection{Analysis} 
The fact that generalization improves as the number of data points per category increases for the PNL model may be explained by the fact that PNL is a specific (supervised) formulation of clustering with Bregman divergences \cite{banerjee2005clustering}. The prototype of each category approximates the point that minimizes the loss in Bregman information \cite{banerjee2005clustering} for that category. 
Bregman information is related to Shannon's rate distortion theory, it corresponds to the optimal distortion-rate to encode a category when the distortion is measured by $d$ (i.e., the squared Euclidean distance). 
The point that minimizes the loss in Bregman information \cite{banerjee2005clustering} for the category is the mean vector of all the examples that belongs to the category. 
Therefore, the larger the number of `shots', the better the approximation of the mean vector of the examples in the category.
However, a larger number of shots does not necessarily lead to better performance as discussed in \cite{snell2017prototypical}.
Statistical guarantees of (a generalization of) PNL are studied in \cite{anonymous2019dimensionality}. In practice, PNL is quite robust to the choice of number of shots. We find that anywhere between 3 to 10 shots work well.
% Without need of choosing sampling strategy and tuning the hyperparameter of margin $\alpha$, PNL models are easier to use. 

% The difference between PNL and TL models is illustrated in Fig.~\ref{fig:compare}. With triplet-based models, each possible triplet in a mini-batch (containing one positive and one negative pair) is considered. When the model is optimized with PNL, a query is compared only with the prototype/centroid of each category. %, but instead of being compared to one negative class, it is compared to all negative classes in the batch.
The reason why SI accuracy of TL models drops dramatically for experiments with more ``ways" might be due to limitation of TL, which has been extensively studied in the literature in different contexts \cite{song2017deep,Law2017}. 
One main limitation is that TL does not necessarily group each category into a single cluster even when the global optimum is reached \cite{song2017deep}. 
%During experiments, we found that with semi-hard sampling strategy, the average loss of triplets in each batch is actually increasing with training.
% This could be explained by the fact that conflicts exist in the optimization directions between triplets. Some triplets will become even "harder" after the model is optimized according to the conflicting triplets. As a result, sampling strategy of triplets should be carefully selected for TL models. Otherwise, it might lead to low converge speed or collapse.

In our experience, PNL is practically easier to use than TL. PNL is not dependent on a triplet sampling strategy, that can impact performance, and do not require the margin parameter $\alpha$. PNL models are also $\sim$3x faster to train (wallclock time) 
than TL models with same mini-batch size because TL requires more pairwise comparisons for batches of same size.%; this however may depend on the details of the implementation.

% \vspace{-1em}
\section{Conclusion}
% \vspace{-.5em}
We have proposed a prototypical network loss-based speaker embedding model, and compared it with the popular triplet loss-based models. 
With identical speech sequence embedding architectures, PNL outperforms the triplet loss when speakers are seen during training, and by an even larger margin on held-out, unseen speakers for both speaker identification and speaker verification tasks. We also illustrate some of the practical advantages of PNL models over TL. In the future, we would like to explore better architectures of speech sequence embedding models and integrate the few-shot learning based speaker embedding model into a speaker diarization pipeline.

%Future work includes comparing models on large scale datasets.

%Training speaker embedding with PNL improves the performance of downstream tasks, such as SV, SI and speaker diarization. 

%\ML{one or two sentences on the possible impact of this method for this task?}

\vfill\pagebreak
\bibliographystyle{IEEEbib}
\bibliography{refs}

\end{document}